\documentclass{article}

\usepackage{PRIMEarxiv}

\usepackage[utf8]{inputenc} 
\usepackage[T1]{fontenc}    
\usepackage{hyperref}       
\usepackage{url}            
\usepackage{booktabs}       
\usepackage{amsfonts}       
\usepackage{nicefrac}       
\usepackage{microtype}      
\usepackage{lipsum}
\usepackage{multirow}
\usepackage{algorithm}
\usepackage{algorithmic}
\usepackage{amsmath}
\usepackage{amsthm}

\newtheorem{Definition}{Definition}
\newtheorem{Theorem}{Theorem}
\usepackage{authblk}
\usepackage{soul}
\usepackage{tabularx}
\usepackage{multirow}

\usepackage{enumerate}
\usepackage{fancyhdr}       
\usepackage{graphicx}       
\graphicspath{{media/}}     
\usepackage[section]{placeins}
\usepackage{caption}
\usepackage{tabularx}
\usepackage[commandnameprefix=always,markup=default,authormarkup=none]{changes}

\title{Knowledge Distillation with Adapted Weight\thanks{This article has been accepted for publication in \href{https://www.tandfonline.com/journals/gsta20}{\textit{Statistics}}, published by Taylor \& Francis}.}

\author[a,b,c]{Sirong Wu}
\author[a,b,c]{Xi Luo}
\author[d,$\dagger$]{Junjie Liu}
\author[a,b,$\dagger$]{Yuhui Deng}
\affil[a]{Guangdong Provincial Key Laboratory of Interdisciplinary Research and Application for Data Science, Zhuhai 519087, China}
\affil[b]{BNU-HKBU United International College, Zhuhai 519087, China}
\affil[c]{Faculty of Science, Hong Kong Baptist University, Hong Kong SAR 999077, China}
\affil[d]{Department of Political Science, Trinity College Dublin, 2 Clare Street, Dublin 2, Ireland}
\affil[$\dagger$]{Correspondence: liuj13@tcd.ie, ivandeng@uic.edu.cn}

\begin{document}
\date{}
\maketitle

\begin{abstract}
Although large models have shown a strong capacity to solve large-scale problems in many areas including natural language and computer vision, their voluminous parameters are hard to deploy in a real-time system due to computational and energy constraints. Addressing this, knowledge distillation through Teacher-Student architecture offers a sustainable pathway to compress the knowledge of large models into more manageable sizes without significantly compromising performance. To enhance the robustness and interpretability of this framework, it is critical to understand how individual training data impact model performance, which is an area that remains underexplored. We propose the \textbf{Knowledge Distillation with Adaptive Influence Weight (KD-AIF)} framework which leverages influence functions from robust statistics to assign weights to training data, grounded in the four key SAFE principles: Sustainability, Accuracy, Fairness, and Explainability. This novel approach not only optimizes distillation but also increases transparency by revealing the significance of different data. The exploration of various update mechanisms within the KD-AIF framework further elucidates its potential to significantly improve learning efficiency and generalization in student models, marking a step toward more explainable and deployable Large Models. KD-AIF is effective in knowledge distillation while also showing exceptional performance in semi-supervised learning with outperforms existing baselines and methods in multiple benchmarks (CIFAR-100, CIFAR-10-4k, SVHN-1k, and GLUE).

\end{abstract}


\section{Introduction}

In the rapidly evolving domain of artificial intelligence (AI), large language models (LLMs) such as GPT-4~\cite{achiam2023gpt} and Gemini~\cite{team2023gemini} have achieved remarkable success in various fields, including Computer Vision~\cite{minaee2021image} and Natural Language Processing~\cite{wei2022emergent}. These models, distinguished by their massive parameter size and intricate architectures, have unlocked unprecedented capabilities, enabling human-like text generation and advanced problem solving. However, deploying such models in real-world scenarios—particularly those with limited computational resources or strict real-time requirements—poses significant challenges. To address this, Hinton et al.~\cite{hinton2015distilling} introduced knowledge distillation, a technique that transfers knowledge from a large, high-performing teacher model to a smaller, more efficient student model. This approach retains much of the teacher's performance while substantially reducing computational demands, making it possible to deploy powerful models on resource-constrained platforms and paving the way for real-time deep learning applications.

Traditional knowledge distillation is typically implemented through an offline learning process~\cite{gou2021knowledge,alkhulaifi2021knowledge}, where a fully trained teacher model guides the training of a compact student model. While this method is straightforward and computationally efficient, it relies on the availability of large-scale datasets and pre-trained teacher models~\cite{zhang2018deep,mirzadeh2020improved,chen2020online}. Furthermore, a higher-performing teacher model does not always guarantee improved performance in the student model and may, in some cases, even degrade the student's performance~\cite{mirzadeh2020improved,cho2019efficacy}. To overcome these limitations, online distillation techniques have been introduced~\cite{zhu2018knowledge,li2021online}, allowing teacher and student models to be trained simultaneously, with the teacher's updates informed by the student's performance~\cite{fan2018learning,zhou2021bert}. However, these methods primarily focus on the training data and do not explicitly address how the student will generalize to unseen data, such as the validation set. Moreover, while these methods facilitate the deployment of powerful models on resource-constrained platforms, they still fall short of addressing key concerns related to safety, interpretability, and robustness—critical factors for real-world AI applications, particularly in sensitive domains such as healthcare and finance ~\cite{giudici2023safe,giudici2024safe}.

\begin{figure}[!ht]
    \centering 
    \includegraphics[width=1\textwidth]{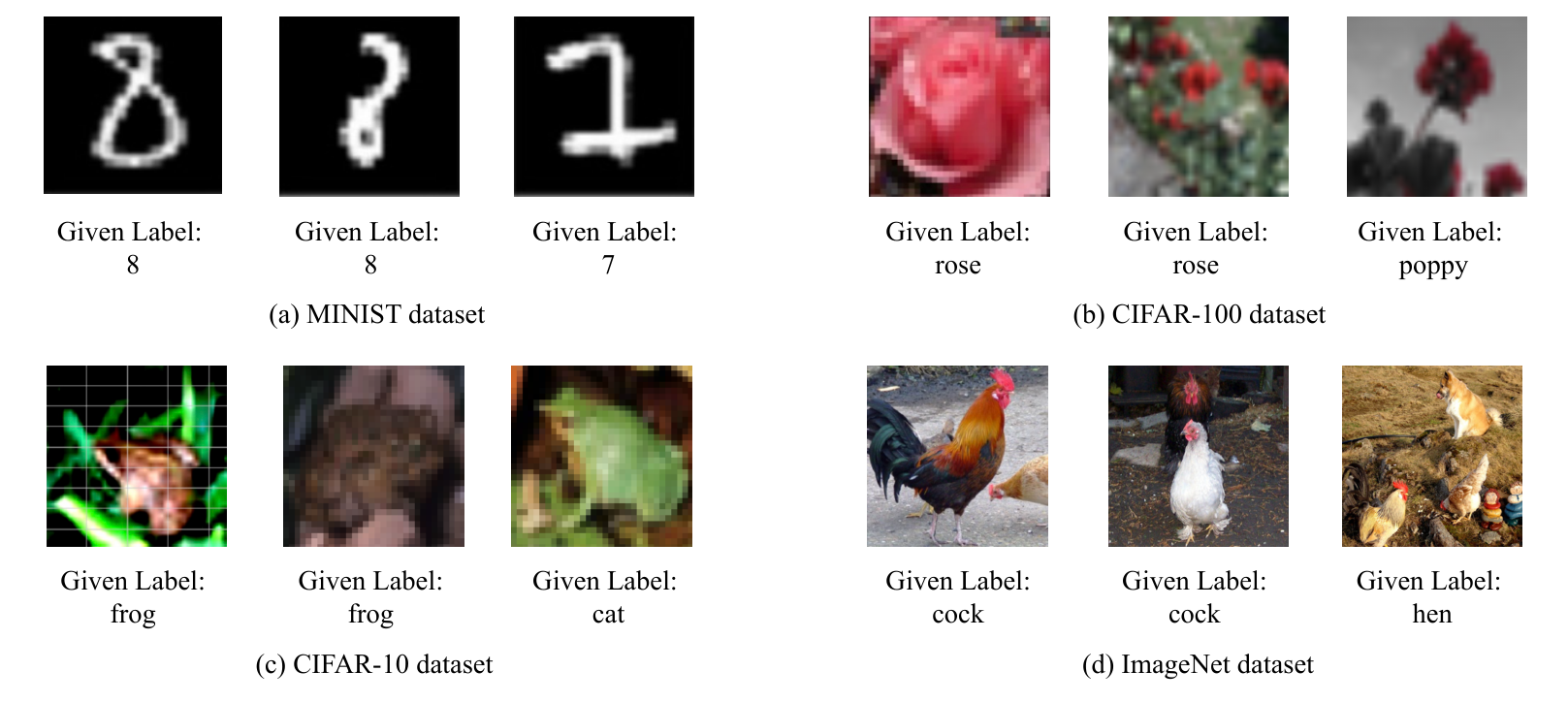}
\caption{Example training data in MINIST, CIFAR-10, CIFAR-100, ImageNet dataset. Left: A helpful training data with the correct label; Mid: A harmful training data with the correct label; Right: A harmful training data with the wrong label. The presence of harmful training data will make the model misclassify the test data.}
\label{fig_1}
\end{figure}

For effective supervised learning, large, high-quality labelled datasets are crucial, requiring extensive human effort for accurate labeling. Current distillation techniques treat all training data equally, assigning a uniform weight to each sample. However, public datasets such as MNIST, CIFAR-10, CIFAR-100, and ImageNet are often riddled with noisy labels, which can mislead the training process and degrade model performance~\cite{zhang2018generalized,li2020dividemix,shu2019meta}, as shown in Figure \ref{fig_1}. For example, a digit in the MNIST dataset may be mislabelled or assigned an incorrect label. Such noise in training data can detrimentally impact generalization performance~\cite{arpit2017closer,zhang2021understanding}. Zhang et al.~\cite{zhang2021understanding} demonstrated that standard Convolutional Neural Networks (CNNs) can overfit noisy labels, resulting in poor generalization. Therefore, applying uniform weights to all training data is not optimal, and assigning individual weights based on data quality and importance becomes necessary.

A promising approach to address this challenge is the influence function, introduced by Koh et al.~\cite{koh2017understanding}, which quantifies the importance of each training data point based on its impact on model performance. By estimating how individual data points affect generalization to unseen data, the influence function provides deeper insights into the learning process, making the model more interpretable and robust. Inspired by this concept, we propose Knowledge Distillation with Adaptive Influence Weight (KD-AIF), which assigns corresponding weights to training data based on their influence on the distillation process. This not only improves the learning efficiency of the student model but also helps the teacher model adapt in a way that maximizes the generalization ability of the student. KD-AIF represents a significant shift from conventional distillation methods, offering a more tailored and explainable approach to model compression.

Moreover, KD-AIF introduces a systematic mechanism that enhances semi-supervised learning, where teacher-student frameworks are also used to leverage unlabelled data more effectively. By integrating the influence of training data into the distillation process, KD-AIF offers a new perspective on how semi-supervised learning can be improved, especially in scenarios with noisy labels. Extensive experiments demonstrate that KD-AIF not only outperforms existing knowledge distillation methods but also exhibits exceptional performance in semi-supervised learning. Following the SAFE principles proposed by Giudici et al.~\cite{giudici2023safe}, which fully account for the Sustainability, Accuracy, Fairness, and Explainability of the model, KD-AIF aligns seamlessly with the SAFE framework in several key ways. Notably, KD-AIF optimizes accuracy by prioritizing data that contributes positively to the student model’s generalization. We validated KD-AIF across multiple domains, showcasing its strong performance and consistent accuracy, and through the use of interpretable influence weights, it adds explainability to the distillation process. Our proposed Influence Weight shares similarities with the "Rank Graduation Robustness" measure (RGR) proposed by Babaei et al.~\cite{babaei2025rank} in assessing robustness, as both methods focus on rank differences in model performance rather than the magnitude of predicted probabilities, making them resilient to outliers and noise. While RGR serves as a statistical metric on the model's evaluation process, KD-AIF focuses on the training process, thereby increasing its applicability to real-world AI systems. In terms of fairness, drawing on the work of Chen et al.~\cite{chen2024measuring}, we intentionally introduce mislabeled data in our experiments to simulate imbalanced and noisy conditions. By addressing biases in training data, it aligns with fairness principles outlined in the SAFE framework. These advancements represent a step toward more transparent, robust, and fair AI systems.

In summary, our contributions are as follows:
\begin{itemize}
    \item We introduce distillation influence weight to measure the impact of each training data point on the student model's generalization ability, thereby enhancing the interpretability of the distillation process.  
    \item We proposed a novel knowledge distillation mechanism: KD-AIF which allow a better instance-level interpretability for deep learning models while enabling a better model performance with limited labelled data. The evaluation is conducted on both classical computer vision and natural language datasets(CIFAR-100, CIFAR-10-4k, SVHN-1k, and GLUE)
\end{itemize}

This paper is structured as follows:  Section \ref{sec:2} formalizes the research question by outlining the problem statement, its significance, and the key objectives addressed in this work. In Section \ref{sec:3}, we present our proposed KD-AIF, detailing the underlying principles, the architecture design, and the implementation steps. Section \ref{sec:4} provides a comprehensive evaluation of the method, showcasing experimental results across multiple metrics to demonstrate its effectiveness and robustness. Section \ref{sec:5} concludes the paper with a discussion of the research implications, highlighting the broader impact of our findings, acknowledging limitations in the current approach, and proposing directions for future work to advance the field.

\section{Background and Problem Statement}\label{sec:2}

\subsection{Assumptions and Base Models}

\paragraph{Assumption}
In most of the machine learning's training process, two datasets are included, training set $\{z^{train}_i \}_{i=1}^{N_{train}}$ and validation set $\{z^{val}_j \}_{j=1}^{N_{val}}$. For each data point $z^{train}_i = (x^{train}_i,y^{train}_i)$, we have $x^{train}_i \in \mathcal{X} \subset \mathbb{R}^{d_x}$, $y \in \mathcal{Y} \subset \mathbb{R}^{d_y}$, where $d_x$ and $d_y$ are the dimension of input and label. Similarly, the validation set can be denoted as $\{z^{val}_j \}_{j=1}^{N_{val}}$, where $N_{val} \ll N_{train}$, and the validation input $x_j^{val} \in \mathcal{X}$. The common assumption of machine learning models is that both $x_i^{train}$ and $x_j^{val}$ are i.i.d and follow the same probability distribution $P(\cdot)$. However, some researchers argue that there is a distribution shift between the training dataset and the validation dataset~\cite{zhang2023revisiting,yang2023change}, i.e.,  $x^{train}_i \in \mathcal{X} \subset \mathbb{R}^{d_x} $ is an i.i.d training input following a probability distribution $P(\cdot)$ while $N_{val} \ll N_{train}$, where the validation input $x^{val}_j \in \mathcal{X}$ follows another probability distribution $Q^\prime(\cdot)$.

In knowledge distillation, the process typically involves two models: a teacher model and a student model. Let $T(x^{train};\theta_T)$ and $S(x^{train};\theta_S)$ respectively be output of the teacher model and student model. The models $T,S \in \mathcal{H}: \mathbb{R}^{d_x} \mapsto\mathbb{R}^{d_y}$ are parameterized by the corresponding model parameters $\theta_T$ and $\theta_S$. $\eta_T$ and $\eta_S$ are the learning rates in training process. The time step is denoted as $m$, which is used to track the evolution of the model parameters during the training process. We denote the cross-entropy loss of two distribution $q$ and $p$ by $L_{ce}(q, p)$. When $q$ represents a label, it is treated as a one-hot distribution. If $q$ and $p$ contain multiple instances (e.g. in a batch), $L_{ce}(q, p)$ represents the average cross-entropy loss computed across all instances in the batch. Specially, we use $L_{kd}(x^{train};\theta_T,\theta_S)$ to represent the loss between the teacher and the student distributions $L_{ce}(T(x^{train};\theta_T), S(x^{train};\theta_S))$.

\paragraph{Vanilla distillation} The vanilla knowledge distillation typically follows a two-stage process. In the first stage, a pre-trained teacher model is fine-tuned to achieve optimal performance on a specific task. In the second stage, a student model is trained to replicate the teacher model’s outputs using the available training data. The optimization objective for the student model is defined as:

\begin{equation}\label{eq:optimize-objective-student}
    L_S(z^{train};\theta_S,\theta_T) 
= (1-\alpha)L_{kd}(x^{train};\theta_T,\theta_S)
+ \alpha L_{ce}(y^{train},S(z^{train};\theta_S))
\end{equation}

where $\alpha$ is the ratio between self-evolution and knowledge transfer. The update of the student follows:

\begin{equation}\label{eq:update-student}
\theta_{S}^{m+1} 
= \theta_{S}^{m} - \eta_S \nabla_{\theta_S} L_S(z^{train};\theta_S,\theta_T)
\end{equation}

The main limitation of the vanilla knowledge distillation is its static treatment of the teacher model. The teacher's parameter remain fixed and ignoring any feedback from the student model once the distillation begins. This rigidity potentially reduces the efficiency and effectiveness of knowledge transfer. However, few existing research works address this limitation by proposing dynamic distillation frameworks. For example, under the vanilla distillation framework, many researchers have explored additional mechanisms to enable the student model to better approximate the performance of the teacher model. Sun et al.~\cite{sun2019patient} introduced PKD mechanism, which allows the student to incrementally extract knowledge by learning from multiple intermediate layers of the teacher model. Mirzadeh et al.~\cite{mirzadeh2020improved} introduced teacher assistant models to act as intermediaries, bridging the performance gap between the teacher and the student. Wang et al.~\cite{wang2022makes} leverage contrastive learning and data augmentation to effectively transfer the teacher model’s representation capabilities to the student, thereby enhancing its performance.

\paragraph{Online distillation}
To enable a distillation process that dynamically adapts to the learning progress of the student model, the concept of online distillation has been introduced. Unlike the vanilla distillation methods, this approach allows for the simultaneous optimization of both the teacher and student models within a single training phrase. In addition to minimizing the cross-entropy loss based on the true labels, online distillation seeks to align the teacher model's output distribution with that of the student model. The models' alignment is achieved by minimizing the knowledge distillation loss, which measures the divergence between the teacher and student models' output distribution. In this way, the teacher model provides guidance that evolves in response to the student's learning state, which fosters a more effective and interactive knowledge transfer process.

\begin{equation}
L_T(z^{train};\theta_T,\theta_S) 
= (1-\alpha)L_{kd}(x^{train};\theta_T,\theta_S)
+ \alpha L_{ce}(y^{train},T(x^{train};\theta_T))
\end{equation}

where $L_{kd}$ is the distillation loss measuring the discrepancy between two models' outputs, and $L_{ce}$ is the cross-entropy loss computed using the true labels. The parameter $\alpha$ balances the contributions of these two objectives.

The training process involves iteratively updating the parameters of both models:
\begin{equation}
\theta_{T}^{m+1} 
= \theta_{T}^{m} - \eta_T \nabla_{\theta_T} L_T(z^{train};\theta_T^{m},\theta_S^{m})
\end{equation}
\begin{equation}
\theta_{S}^{m+1} 
= \theta_{S}^{m} - \eta_S \nabla_{\theta_S} L_S(z^{train};\theta_S^{m},\theta_T^{m+1})
\end{equation}

By engaging in a process of continuous feedback and updates, the student model benefits from observing the teacher model's learning trajectory, enhancing its ability to perform the specified task more effectively. Building on this framework, several studies have proposed more efficient online distillation models. For instance, Similarity-Preserving model (SP) proposed by Tung and  Mori~\cite{tung2019similarity} is an online knowledge distillation method that facilitates knowledge transfer by maintaining the relative similarity between samples in the feature space. Amara et al.~\cite{amara2022bd} introduced BD-KD, an adaptive method that dynamically balances reverse and forward divergences during the online distillation process, enhancing the efficiency and robustness of knowledge transfer.

Nonetheless, the primary emphasis of online distillation on knowledge transfer from teacher to student within the confines of the training dataset does not inherently ensure the student model's adeptness on novel, unseen data. This oversight may culminate in the student model's proficiency being confined to the training data, without adequately developing the capacity to generalize to new situations.

\subsection{Problem Statement}
Based on the i.i.d assumption that we mentioned in the beginning of this section, for the standard training process of online distillation, we aim to minimize the empirical risks $R_{\theta_T}(P)$ and $R_{\theta_S}(P)$, and learn the optimal parameters $\hat{\theta}_T$ and $\hat{\theta}_S$.
\begin{equation}
\hat{\theta}_T = \underset{\theta_T}{argmin}\,R_{\theta_T}(P) = \underset{\theta_T}{argmin}\,\mathbb{E}_{P}[L_T(z^{train};\theta_T,\theta_S)]
\end{equation}
\begin{equation}
\hat{\theta}_S = \underset{\theta_S}{argmin}\,  R_{\theta_S}(P) =  \underset{\theta_S}{argmin}\,\mathbb{E}_{P}[L_S(z^{train};\theta_T,\theta_S)] 
\end{equation}
Within the teacher-student framework, our primary focus is on the student model's performance and improvement. Specifically, we aim to evaluate whether the student effectively leverages the teacher's guidance to achieve enhanced accuracy, generalization, and efficiency. In other words, we are particularly concerned with the performance of the student model on the unseen data (validation set) and to ensure the generalization ability, we adopt the $OOD$ assumption instead of the i.i.d assumption in the student's objective function, denoted as $R_{\hat{\theta}_S}(Q^\prime)$. It is worth noting that when calculating the validation risk $R_{\hat{\theta}_S}(Q^\prime)$, $\theta_T$ is already fixed.
\begin{equation}
R_{\hat{\theta}_S}(Q^\prime) = \mathbb{E}_{Q^\prime}[L_S(z^{val};\theta_T,\hat{\theta}_S)]
\end{equation}
The primary challenge arises from the fact that $\hat{\theta}_S$, the student model parameters learned during training, may not be the optimal risk minimizer for the validation distribution (or test distribution). This is due to two main factors: the distribution shift between the training distribution $P$ and the validation distribution $Q^\prime$, and the presence of noisy or mislabelled data in the training set. Such discrepancies can lead to suboptimal performance when the model is applied to the validation or test distribution.

To address this problem, our goal is to learn appropriate weights for the input data, enabling the model to prioritize informative data while reducing the influence of harmful or noisy ones. By assigning lower weights to detrimental training data, we effectively minimize their adverse impact on the learning process. This reweighting strategy can result in an updated model that achieves a lower test risk, namely $R_{\tilde{\theta}_S}(Q^\prime) < R_{\hat{\theta}_S}(Q^\prime)$, where $\tilde{\theta}_S$ represents the updated student model parameter learned after incorporating weights into the training process. 




\section{Methods}\label{sec:3}

\subsection{Influence Function}
Highly complex and over-parameterized models, such as deep neural networks, are often regarded as "black boxes" due to the challenge of interpreting their internal mechanisms and decision-making processes. The impact of individual training data on these models is substantial; even the addition or removal of a single training instance may lead to significant changes in the model's behaviour. In extreme cases, accurately assessing the impact of a single data point on the model's performance might necessitate retraining the entire model—a computationally expensive and often impractical task. This lack of transparency limits our understanding of how these models generalize to unseen data and what specific training instances contribute to their predictions.

To tackle this issue, the concept of the influence function was introduced to provide a way to interpret these "black-box" models~\cite{koh2017understanding}. Influence function measures how individual training data affects the model's predictions by estimating the impact of each instance on the model's learned parameters. This approach provides a quantitative framework for understanding how specific training data influence the model's overall performance. It is particularly useful for identifying outliers or mislabelled data that disproportionately affect the model's behavior. By calculating the influence function for each data, one can predict how modifying, removing, or emphasizing that data would alter the model's predictions, offering insights into its decision-making process.

More specifically, the influence function computes the change of the optimal model parameters when upweighting the loss of a training data or perturbing a training data, therefore, the influence function can help us identify which training data $z_i^{train}$ is most responsible for the prediction of unseen data. In this paper, we use the concept of influence function to construct influence weights, which estimates how much will the student’s performance on unseen data change if we remove one training data in the knowledge distillation process. For each training data $z_i^{train}$, we compute its influence based on validation data instead of test data since test data should be invisible until testing phase. The key idea is to make a first-order Taylor approximation of the changes in $\hat{\theta}_S$ around $\epsilon =0$. Hence, the influence is rewritten as follows.



Let $\hat{\theta}_S = \underset{\theta_S}{argmin} \frac{1}{N_{train}}\sum^{N_{train}}_{i=1} L_{S}(z_i^{train};\theta_S,\theta_T)$ denote the optimal parameters of the student model. If the $i^\prime$-th training sample is upweighted by a small $\epsilon_{i^\prime}$, then the perturbed optimal student model parameter $\hat{\theta}_{S,\epsilon_{i^\prime}}$ becomes 
\begin{equation}
\hat{\theta}_{S,\epsilon_{i^\prime}}
= \underset{\theta_S}{argmin} \frac{1}{N_{train}}\sum^{N_{train}}_{i=1} L_{S}(z_i^{train};\theta_S,\theta_T) + \epsilon_{i^\prime} 
L_{S}(z_{i^\prime}^{train};\theta_S,\theta_T)
\end{equation}

Based on the derivation of Influence Function~\cite{koh2017understanding}, the change of model parameters is 

\begin{equation}
\hat{\theta}_{S,\epsilon_{i^\prime}} - \hat{\theta}_S 
\approx  \frac{d\hat{\theta}_{S,\epsilon_{i^\prime}}}{d\epsilon_{i^\prime}}\bigg|_{\epsilon_{i^\prime}=0} \epsilon_{i^\prime} 
= - H_{\hat{\theta}_S}^{-1} \nabla_{\theta_S} L_{S}(z_{i^\prime}^{train};\hat{\theta}_S,\theta_T) \epsilon_{i^\prime} 
\end{equation}

where $H_{\hat{\theta}_S} \overset{def}{=} \frac{1}{N_{train}}\sum^{N_{train}}_{i=1} \nabla_{\theta}^{2} L_{S}(z_{i}^{train};\hat{\theta}_S,\theta_T)$ is the Hessian matrix.

Similarly, the influence of upweighting $z_i^{train}$ by a small $\epsilon_i$ on the loss at validation data can be written as  

\begin{equation}\label{eq:I_definition}
\begin{aligned}
\mathcal{I}(z_i^{train},z_j^{val};\hat{\theta}_S)
&= \frac{d(L_{S}(z_{j}^{val};\hat{\theta}_{S,\epsilon_i},\theta_T) -L_{S}(z_{j}^{val};\hat{\theta}_S,\theta_T))}{d\epsilon_i}\bigg|_{\epsilon_i=0} 
= \frac{d L_{S}(z_{j}^{val};\hat{\theta}_{S,\epsilon_i},\theta_T)}{d\epsilon_i}\bigg|_{\epsilon_i=0} \\
&= - \nabla_{\theta_S} L_{S}(z_{j}^{val};\hat{\theta}_S,\theta_T)^{T} H_{\hat{\theta}_S}^{-1} \nabla_{\theta_S} L_{S}(z_{i}^{train};\hat{\theta}_S,\theta_T)
\end{aligned}
\end{equation}

Once $\mathcal{I}(z_i^{train},z_j^{val};\hat{\theta}_S)$ is computed, we will be able to compute the average influence score $\phi_i$ for each training data $z_i^{train}$ across $N_{val}$ validation data.

\begin{equation}\label{eq:influence_i}
\phi_i(\hat{\theta}_S) = \frac{1}{N_{val}} \sum_{j=1}^{N_{val}} \mathcal{I}(z_i^{train},z_j^{val}) = -\frac{1}{N_{val}} \sum_{j=1}^{N_{val}} \nabla_{\theta_S} L_{S}(z_{j}^{val};\hat{\theta}_S,\theta_T)^{T} H_{\hat{\theta}_S}^{-1} \nabla_{\theta_S} L_{S}(z_{i}^{train};\hat{\theta}_S,\theta_T)
\end{equation}

This approach allows for a more nuanced and precise optimization of the knowledge distillation process, ensuring that each data point's contribution is tailored to maximize the student model's generalization capabilities.

\subsection{Influence Weight}

Consider there are $N_{train}$ perturbations $\{\epsilon_i \}_{i=1}^{N_{train}}$ on each training data $x_i^{train}$, let $\mathbf{\epsilon} = (\epsilon_1, \epsilon_2, \dots, \epsilon_{N_{train}})^{T} $, the perturbed risk minimizer of the student model is denoted as $\hat{\theta}_{S,\epsilon}$. 
\begin{equation}\label{eq:preturbed_train_risk}
    \hat{\theta}_{S,\epsilon}
    =\underset{\theta_S}{argmin}\, R_{\hat{\theta}_{S,\epsilon}}(P)
    = \underset{\theta_S}{argmin}\, \mathbb{E}_{P} [L_{S}(z^{train};\theta_S,\theta_T)] + \mathbf{\epsilon}^{T} \
    L_{S}(z^{train};\theta_S,\theta_T)
\end{equation}

Given the validation data from another distribution $Q^\prime$, the objective is to design $\mathbf{\epsilon}$ that minimizes the validation risk $R_{\hat{\theta}_{S,\epsilon}}(Q^\prime)$. According to the definition of Influence Function in Equation \ref{eq:I_definition}, if $z_i^{train} \sim P$ is upweighted by $\epsilon_i$, the loss change of $z_j^{val} \sim Q^\prime$ can be approximated as:
\begin{equation}
    L_{S}(z_{j}^{val};\hat{\theta}_{S,\epsilon_i},\theta_T) - L_{S}(z_{j}^{val};\hat{\theta}_{S},\theta_T) \approx \epsilon_i \times \mathcal{I}(z_i^{train},z_j^{val};\hat{\theta}_S)
\end{equation}

It can be extended to the whole validation distribution:
\begin{equation}
    R_{\hat{\theta}_{S,\epsilon_i}}(Q^\prime) - R_{\hat{\theta}_{S}}(Q^\prime) 
    = \mathbb{E}_{Q^\prime}[ L_{S}(z^{val};\hat{\theta}_{S,\epsilon_i},\theta_T) - L_{S}(z^{val};\hat{\theta}_{S},\theta_T)]
    \approx \epsilon_i \times \mathbb{E}_{Q^\prime}[\mathcal{I}(z_i^{train},z^{val};\hat{\theta}_S)]
\end{equation}

We assume the perturbations $\{\epsilon_i \}_{i=1}^{N_{train}}$ on each training data are small and each training data influences the validation risk independently. Therefore, let $\Phi(\hat{\theta}_S) = (\phi_1(\hat{\theta}_S), \phi_2(\hat{\theta}_S), \dots, \phi_{N_{train}}(\hat{\theta}_S))^{T} $, using the notation in Equation \ref{eq:influence_i}, the validation risk change can be approximated as:
\begin{equation}\label{eq:whole_valid_risk}
    R_{\hat{\theta}_{S,\epsilon}}(Q^\prime) - R_{\hat{\theta}_{S}}(Q^\prime) 
    \approx \mathbb{E}_{P}[\epsilon_i \times \mathbb{E}_{Q^\prime}[\mathcal{I}(z_i^{train},z^{val};\hat{\theta}_S)]]
    = \mathbb{E}_{P}[\mathbf{\epsilon}^{T} \, \Phi(\hat{\theta}_S) ]
\end{equation}


\begin{Theorem}\label{theorem1}
    The performance of the perturbed model $\hat{\theta}_{S,\epsilon}$ on the validation distribution $Q^\prime$ will be better than that of the original model $ \hat{\theta}_{S}$, if the following two conditions are satisfied: 1. $\epsilon_i = 0$ when $\phi_i(\hat{\theta}) = 0$; 2. The perturbation $\epsilon_i$ is a decreasing function $f_{\epsilon}(\cdot)$ of the influence score $\phi_i(\hat{\theta})$.
\end{Theorem} 
The detailed proof of Theorem \ref{theorem1} can be found in Appendix A

According to Equation \ref{eq:whole_valid_risk}, minimizing the validation risk is equivalent to minimizing the function $\mathbb{E}_{P}[\mathbf{\epsilon}^{T} \, \Phi(\hat{\theta}_S) ]$. Theorem \ref{theorem1} gives the instruction that as long as the perturbation $\{\epsilon_i \}_{i=1}^{N_{train}}$ satisfies two conditions related to the influence function, the perturbed model will achieve a lower validation risk compared to the original model. 

In the training process, the perturbed training risk in Equation \ref{eq:preturbed_train_risk} is defined in the perturbation $\epsilon_i$ instead of the influence weight $w_i$, so that we need to bridge the gap between them:
\begin{equation}
\begin{aligned}
    R_{\hat{\theta}_{S,\epsilon}}(P) 
    & \approx \frac{1}{N_{train}}\sum^{N_{train}}_{i=1} L_{S}(z_i^{train};\theta_S,\theta_T) + \sum^{N_{train}}_{i=1} \epsilon_{i} L_{S}(z_{i}^{train};\theta_S,\theta_T) \\
    & = \frac{1}{N_{train}}\sum^{N_{train}}_{i=1} (N_{train} \times \epsilon_{i} +1)
    L_{S}(z_i^{train};\theta_S,\theta_T)
\end{aligned}
\end{equation}
Here we assume $\epsilon_i \in [-\frac{1}{N_{train}},\frac{1}{N_{train}}]$.  $\epsilon_i =0$ means no perturbation is applied, while $\epsilon_i =-\frac{1}{N_{train}}$ means that training data are completely dropped in the training object function. Since the perturbation $\epsilon_i$ is the function of the influence score $\phi_i(\hat{\theta}_S)$, therefore, we define the weight function as follows,
\begin{equation}
    w_i = f_w(\phi_i(\hat{\theta}_S)) = N_{train} \times f_{\epsilon}(\phi_i(\hat{\theta}_S)) +1 \in [0,2]
\end{equation}
where $f_{w}(\cdot) \in \mathcal{F}:\mathbb{R} \to \mathbb{R}$ , then the perturbation $\{\epsilon_i \}_{i=1}^{N_{train}}$ is transformed to the influence weight $\{w_i \}_{i=1}^{N_{train}}$.

\subsection{Analysis of weight function}

The influence weight $w_i$ is calculated based on a particular validation set $Q^\prime$, making it crucial to discuss how different validation sets can affect the calculated influence score. The variation in validation data can lead to different influence weights, as the model's performance on different subsets of the data can vary significantly. Understanding this variability is key to evaluating how robust the influence measure is to changes in the validation set and to ensure that the calculated influence accurately reflects the model’s generalization ability across different data distributions.

Consider the unseen dataset $ \mathcal{Q}= \{ Q \mid Q \ll P, D_{\chi^2}(Q \parallel P) \leq \delta, \delta \geq 0 \} $, where $Q \ll P $ indicates that $Q $ is absolutely continuous with respect to  $P$ , and $ D_{\chi^2}(\cdot \parallel \cdot)$ represents the $\chi^2$ -divergence. The set $\mathcal{Q}$ defines a $ \chi^2 $-divergence neighborhood around the training data distribution $ P $, characterizing all unseen data distributions that are "close" to $ P $ in terms of the $ \chi^2 $-divergence. The worst-case risk $ \acute{R}_{\hat{\theta}_\epsilon}(Q) $ is defined as the supremum of the risk over all $ Q \in \mathcal{Q} $ ~\cite{bagnell2005robust}:
\begin{equation}\label{eq:supremum_risk}
    \acute{R}_{\hat{\theta}_{S,\epsilon}}(Q) = \sup_{Q \in \mathcal{Q}} \mathbb{E}_{Q} \left[ L_{S}(z^{val};\hat{\theta}_{S,\epsilon},\theta_T)  \right]
\end{equation}

In a more convenient form, as derived by Duchi et al.~\cite{duchi2021learning}, the dual representation of this equation is given by:
\begin{equation}\label{eq:dual_risk}
\acute{R}_{\hat{\theta}_{S,\epsilon}}(\eta) = \inf_{\eta \in \mathbb{R}} \left( \sqrt{2\delta + 1} \times \mathbb{E}_P \left[ (L_{S}(z^{train};\hat{\theta}_{S,\epsilon},\theta_T) - \eta)_+^2 \right]^{\frac{1}{2}} + \eta \right)
\end{equation}
where $ \eta $ represents the dual variable. This duality allows us to transform the supremum in Equation \ref{eq:supremum_risk} into a convex function, enabling a more practical and quantitative measurement of the worst-case risk over training data distribution $ P $.

Before analyzing how the worst-case risk $ R_{\hat{\theta}_\epsilon}(Q) $ changes under different weight functions, as discussed in Theorem 2, it is important to define the following concepts:
\begin{Definition}
    A function $ f(x) : \mathbb{R}^d \to \mathbb{R} $ is Lipschitz continuous with constant $ \xi $ if $\| \nabla f(x) - \nabla f(y) \| \leq \xi \| x - y \|$, for all $x, y \in \mathbb{R}^d$.
\end{Definition}
\begin{Definition}
    A function $ f(x) $ has gradients that are bounded by $ \sigma $ if  $\| \nabla f(x) \| \leq \sigma$, for all $x \in \mathbb{R}^d$.
\end{Definition}

These definitions will help us characterize the behavior of the worst-case risk and its dependency on the weight functions used.
\begin{Theorem}\label{theorem2}
    Let $ \hat{\eta} $ be the optimal dual variable that minimizes the expression in Equation \ref{eq:dual_risk}, and assume that the influence weight function $ f_w( \phi(\hat{\theta}_S)) $ has gradients that are bounded by $ \sigma $. Under these conditions, the worst-case risk $ R_{\hat{\theta}_\epsilon}(\hat{\eta}) $ is a Lipschitz continuous function with respect to the influence function vector $ \Phi(\hat{\theta}_S) $. The Lipschitz constant for this function is given by $\mathcal{O}( \frac{\sigma\sqrt{2\delta + 1}}{N_{train}}) $, the gradient of the worst-case risk with respect to $ \Phi(\hat{\theta}_S) $ is bounded by:
\begin{equation}
    \|\nabla_{\boldsymbol{\phi}} \acute{R}_{\hat{\theta}_{S,\epsilon}}(\hat{\eta})\| \leq  \frac{\sigma\sqrt{2\delta + 1}}{N_{train}} \times \|\Phi(\hat{\theta}_S)\|
\end{equation}
\end{Theorem}
Theorem 2 establishes a connection between the rate of change of the worst-case risk and the gradient bound $\sigma$ of the weight function $f_w$. This theorem highlights that setting the weights of certain data points—those deemed less informative or detrimental—to zero can cause the worst-case risk to lose its Lipschitz continuity. Specifically, the inconsistency at zero can result in unbounded gradients, leading to $\sigma \to  \infty$, which in turn introduces severe fluctuations in the risk. At the same time, Theorem 2 provides practical guidance on selecting an appropriate weight function. By tuning $\sigma$ properly, the weighting strategy can achieve improved generalization performance, balancing robustness and stability. This underscores the importance of carefully adjusting the weight function to avoid instability while enhancing out-of-sample performance. The detailed proof of Theorem\ref{theorem2} can be found in Appendix B.

\subsection{KD-AIF}

As shown above, the influence weight derived from the perturbation reflects the importance of each data point in contributing to the model’s generalization ability. Specifically, data that positively impact the student model’s ability to generalize are those for which the gradient of the training loss is aligned with the gradient of the validation loss. In other words, the influence function value is lower for such data points, which corresponds to assigning them higher weights. This adaptive weighting mechanism ensures that the most beneficial data for generalization are prioritized in the learning process, while less useful or misleading data are downweighted.

According to Theorem \ref{theorem1}, the influence weight must be a decreasing function of the influence score and should be evaluated to 0 when the influence score is 0. Furthermore, as stated in Theorem 2, the gradient upper bound of the weight function should be carefully controlled to ensure stable generalization performance. To meet these criteria, the Sigmoid function can be employed to transform the normalized influence score $\tilde{\phi}_i(\hat{\theta}_S)$ into weights, ensuring compliance with these requirements.
\begin{equation}
    w_i = f_w(\tilde{\phi}_i(\theta_S)) 
    = \frac{2}{1+e^{\tilde{\phi}_{i}(\theta_S)}}
\end{equation}

Under the framework of knowledge distillation, notice that the optimal parameter $\hat{\theta}_S$ always depends on the teacher parameter $\theta_T$ via the soft label $T(z^{train};\theta_T)$, hence we can explicitly express the dependency as $\theta^{\prime}_{S} (\theta_T)$. Let $R_{\theta}(P)$ denotes the empirical risk in training process, it is clearly seen that $R_{\theta^{\prime}_{S}(\theta_T)}(P)$ is also a function of $\theta_T$. We use $w_i (\theta)$ to represent the influence weight $w_i$ calculated under the model parameter $\theta$,  then we propose to train the teacher model using the following objective:

\begin{equation}
\hat{\theta}_T = 
\mathop{argmin}\limits_{\theta_T} \, R_{\theta^{\prime}_{S}(\theta_T)}(P) = \mathop{argmin}\limits_{\theta_T} \, \sum_{i=1}^{N_{train}} w_i (\theta^{\prime}_{S} (\theta_T)) L_T(z_{i}^{train};\theta_T,\theta^{\prime}_{S}(\theta_T))
\end{equation}
where 
\begin{equation}
\theta^{\prime}_{S} (\theta_T) 
= \mathop{argmin}\limits_{\theta_S} R_{\theta_S}(P) 
= \mathop{argmin}\limits_{\theta_S} \sum_{i=1}^{train} w_i (\theta_S)  L_S(z_{i}^{N_{train}};\theta_T,\theta_S)
\end{equation}

\begin{figure}[!ht]
    \centering
    \includegraphics[width=1\textwidth]{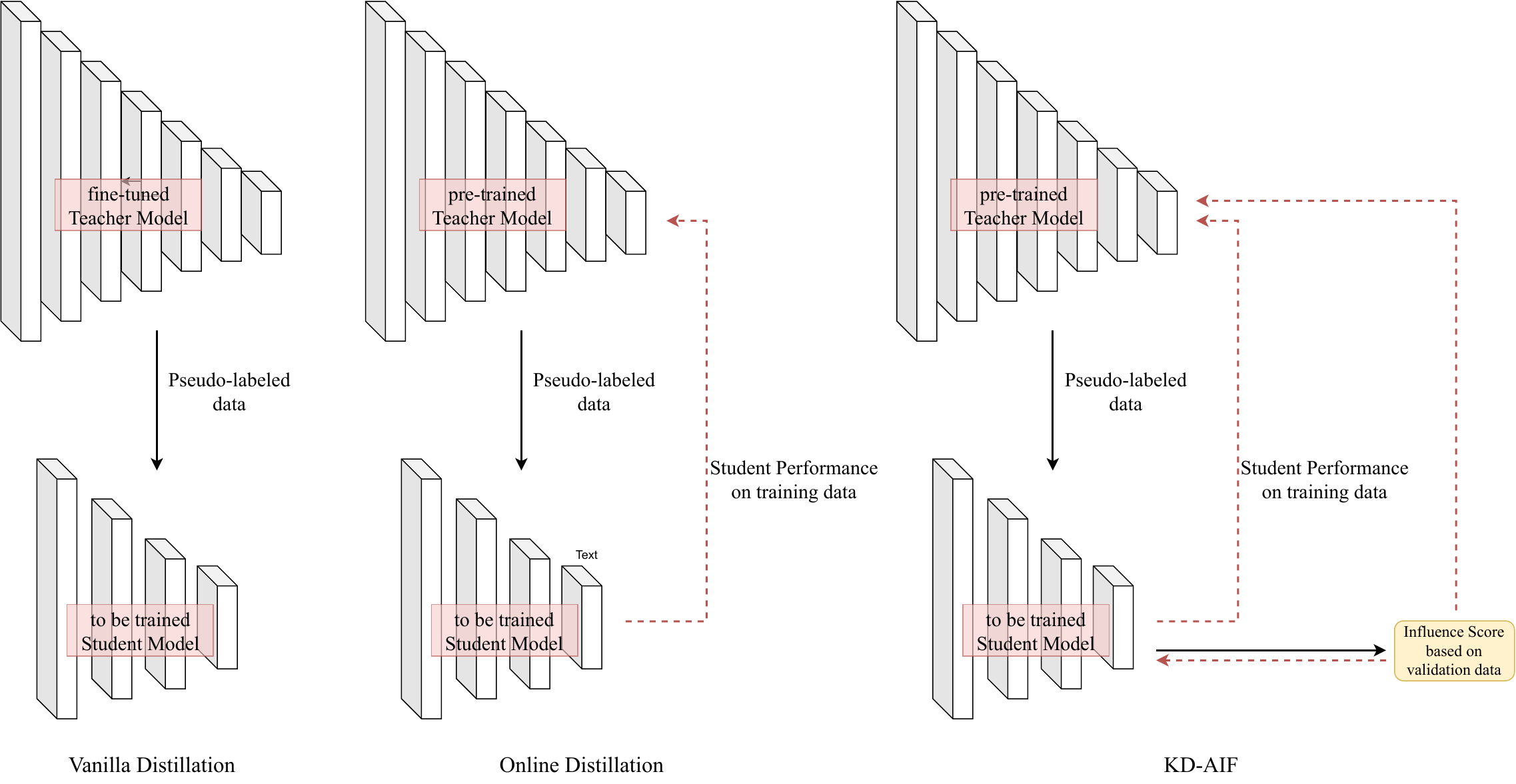}
    \caption{Comparison of vanilla distillation, online distillation, and our proposed framework KD-AIF. The dotted red lines show the feedback from student model. Note that vanilla distillation employs a two-stage training pipeline by first fine-tuning the teacher model on the target task. Both online distillation and KD-AIF adopt a one-stage joint training strategy, allowing simultaneous training of the teacher and student models. Unlike online distillation's equal weighting, KD-AIF introduces an adaptive feedback mechanism based on influence scores, which evaluates the importance of training data and dynamically adjusts the training process to improve generalization and robustness.}
    \label{fig_2}
\end{figure}

Overall, our method allows the teacher to adapt to the student’s generalization abilities and provide more personalized guidance to student. We illustrate the general workflow of vanilla distillation, online distillation and KD-AIF in Figure \ref{fig_2}. It can be clearly seen from Figure 2 that in the vanilla distillation model, the teacher model remains static and is not updated after training. In traditional online distillation models, the teacher and student models learn from each other dynamically, but these approaches do not explicitly account for generalization performance on unseen data. This limitation is particularly evident when the validation (or test) data distribution deviates significantly from the training data, potentially leading to poor transferability. Our proposed KD-AIF model addresses these challenges by incorporating the student model's generalization performance on validation data into the mutual learning process. By feeding this feedback into the interaction between the teacher and student models, KD-AIF ensures that both models improve simultaneously while enhancing their generalization capability. This approach fosters better alignment and performance on unseen data, addressing a critical gap in traditional distillation frameworks. Additionally, the detailed KD-AIF framework is outlined in algorithm \ref{alg3:AIF}.

\begin{algorithm}
	\caption{KD-AIF}
	\label{alg3:AIF}
	\begin{algorithmic}[1]
		\REQUIRE Initialize $\theta_T^{(0)}$ and $\theta_S^{(0)}$, learning rate $\eta_T$ and $\eta_S$; \\
        $N_{train}$, $N_{val}$ represent the total number of data in train dataset and validation dataset, respectively;\\
        the maximum number of the training steps $M$
        \ENSURE $\theta_T^{(M)}$ and $\theta_S^{(M)}$
        \FOR{$t=0$ to $M-1$}
            \FOR{$i=1$ to $N_{train}$}
            \FOR{$j=1$ to $N_{val}$}
		\STATE Calculate influence score $\mathcal{I}(z_i^{train},z_j^{val};\theta_S^{(0)})$ for each training data on a particular validation data $z_{j}^{val}$
        \ENDFOR 
        \STATE Calculate the average influence score $\phi_i(\theta_S^{(0)})$ for each training data $z_i$
        \ENDFOR 
        \STATE Use normalized influence score $\Tilde{\phi}_i$ to calculate distillation influence weight $w_i$ for each training data
        \STATE Update teacher $\theta_T$: $\theta_T^{(1)} \leftarrow \theta_T^{(0)} - \eta_T \nabla_{\theta_T} \sum_i^{N_{train}} w_i L_T(z_{i}^{N_{train}};\theta_T^{(0)},\theta_S^{(0)})$
        \STATE Update student $\theta_S$: $\theta_S^{(1)} \leftarrow \theta_S^{(0)} - \eta_S \nabla_{\theta_S} \sum_i^{N_{train}} w_i L_S(z_{i}^{N_{train}};\theta_T^{(1)},\theta_S^{(0)})$
		\ENDFOR 
	\end{algorithmic}
\end{algorithm}

To analyse the Algorithm \ref{alg3:AIF}'s efficiency, we adopt the algorithm time complexity as an evaluation. The dominant operations are within the nested loops, specifically the calculation of the influence score $\mathcal{I}(z^{train}, z^{val};\theta_{S})$. This calculation is performed $M \cdot N_{train} \cdot N_{val}$ times for the entire nested loops, i.e., $O(M \cdot N_{train} \cdot (N_{val} + P_T + P_S))$, where $P_T$ and $P_S$ represent the number of teacher and student model's parameter, respectively, and commonly, the backpropogation process is linear. For $\mathcal{I}(\cdot, \cdot;\theta_{S})$, the most computationally expensive part of the equation is the calculation and inversion of the Hessian matrix. The Hessian matrix has dimensions equal to the number of student parameters squared. Calculating all the second-order partial derivatives requires $O(P_S^2)$ operations, and the inversion of the Hessian is generally $O(P_S^3)$ with standard methods like Gaussian elimination. Thus, the overall complexity is $O(M \cdot N_{train} \cdot (N_{val} \cdot P_S^3 + P_T + P_S))$. In practice, the Hessian matrix and its inversion complexity can be reduced, for example, the EK-FAC~\cite{martens2015optimizing, george2018fast, eschenhagen2024kronecker} and L-BFGS~\cite{berahas2016multi} algorithm can help reduce it from $O(P_S^3)$ to $O(P_S^2)$. Hence, the computational complexity of KD-AIF is $O(M \cdot N_{train} \cdot (N_{val} \cdot P_S^2 + P_T + P_S))$.

\subsection{KD-AIF in Semi-Supervised Learning}

In semi-supervised learning, Pseudo Label methods ~\cite{arazo2020pseudo,pham2021meta} have significantly advanced state-of-the-art models across various fields. These methods employ a teacher-student network framework, where the teacher model generates pseudo labels for unlabelled data. The student model then trains on both the pseudo-labelled and labelled data, leveraging the expanded dataset and regularization techniques such as data augmentation. Tarvainen and Valpola\cite{tarvainen2017mean} proposed an improved semi-supervised learning method by enhancing the teacher-student framework with data augmentation and consistency regularization. Li et al.~\cite{li2024dsst} introduced a dual-student model that eliminates the constraints of a teacher model, enabling the two student models to learn collaboratively, which significantly improves semi-supervised learning performance. Additionally, Pham et al.~\cite{pham2021meta} proposed Meta Distillation, a framework employing two relatively complex models in a teacher-student setup to refine pseudo-labels and distillation loss, thereby improving performance on unlabeled data effectively. In semi-supervised learning, the student model is often as complex as the teacher model, unlike in distillation frameworks where the student is typically simpler.

However, a major drawback is that if the pseudo labels are inaccurate, the student may learn from flawed data, leading to limited improvement over the teacher. It is natural to believe that not all the training data including labelled and unlabelled data are equal. Despite this, our approach demonstrates that even using a lightweight student model, our mechanism consistently achieves significant improvements. Experiments validate that under this framework, the results surpass many existing methods, showcasing the robustness and effectiveness of our design in leveraging lightweight student models.  

\begin{figure}[!ht]
    \centering
    \includegraphics[width=0.8\textwidth]{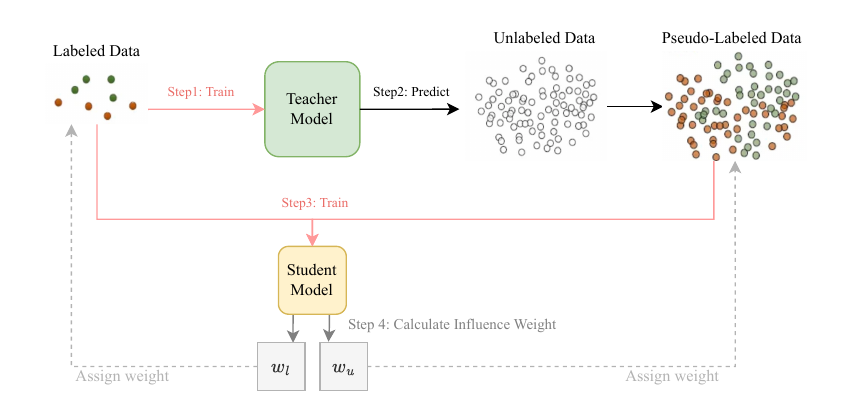}
    \caption{The workflow of KD-AIF framework in Semi-Supervised Learning. The teacher model is initially trained on labelled data (Step 1) and then used to generate pseudo-labels for unlabelled data (Step 2). The labelled and pseudo-labelled data are then combined to train} the student model (Step 3). In Step 4, influence weights are calculated for both labelled and pseudo-labelled data, assigning adaptive weights to these datasets based on their importance. The teacher and student models are further updated based on these weighted datasets, and the process is iteratively repeated to refine the performance of both models.
    \label{fig_3}
\end{figure}

We present the workflow of KD-AIF in Semi-Supervised Learning as shown in Figure \ref{fig_3}.The training process is structured into iterative steps involving both the teacher and student models. Initially, the teacher model is trained using labelled data to capture essential patterns (Step 1). Following this, the trained teacher model generates pseudo-labels for unlabelled data, creating a pseudo-labelled dataset (Step 2). This combined dataset of labelled and pseudo-labelled data is then used to train the student model, allowing it to leverage the expanded data volume (Step 3). In the subsequent step (Step 4), influence weights are calculated for both the labelled and pseudo-labelled datasets to assign adaptive weights that emphasize their relative importance. These influence weights guide the further updating of both the teacher and student models, ensuring that the training process focuses on more informative samples. This iterative procedure is repeated, progressively refining the teacher and student models for improved performance and robust generalization.

Let $\{z_l^{train} \}_{l =1}^{N_{l}}$ be the labelled training set, $z_l^{train} = (x^{train}_l,y^{train}_l)$, $\{z_u^{train} \}_{u =1}^{N_{u}}$ be the unlabelled training set, $z^{train}_u = (x^{train}_u,y^{train}_u)$. Originally, the student model is trained based on the pseudo-labelled data generated by the teacher model $T(z_u;\theta_T)$ with labelled data. The teacher model is trained based on the labelled data solely. By incorporating the influence weight, the semi-supervised optimization problem can be further formulated as follows.
\begin{equation}
    \hat{\theta}_T
= \mathop{argmin}\limits_{\theta_T} \sum_{l=1}^{N_{l}} w_l(\theta^{\prime}_{S} (\theta_T)) L_{ce}(y_l^{train},T(x_l^{train};\theta_T)
\end{equation}
where 
\begin{equation}
\theta^{\prime}_{S} (\theta_T) 
=\mathop{argmin}\limits_{\theta_S} \left[ \sum_{l=1}^{N_{l}} w_l L_{ce}(y_l^{train},S(x_l^{train};\theta_S))  + \alpha \sum_{u=1}^{N_{u}} w_u (\theta_S) L_{kd}(x_u^{train};\theta_T,\theta_S) \right]
\end{equation}
\begin{equation}
     w_u = f_w(\tilde{\phi}_u(\theta_S)),\quad \phi_u(\theta_S) = \frac{1}{N_{val}} \sum_{j=1}^{N_{val}} \mathcal{I}(z_u^{train},z_j^{val})
\end{equation}
\begin{equation}
     w_l = f_w(\tilde{\phi}_l(\theta_S)),\quad \phi_l(\theta_S) = \frac{1}{N_{val}} \sum_{j=1}^{N_{val}} \mathcal{I}(z_l^{train},z_j^{val})
\end{equation}

\section{Experiments}\label{sec:4}

The experiments were conducted on a setup equipped with an NVIDIA RTX 4090 GPU (24GB RAM), 16GB system RAM, and an Intel i7 processor. The operating system is Ubuntu 22.04, CUDA\footnote{\href{https://developer.nvidia.com/cuda-toolkit}{CUDA Tool-Kit}} version is 11.7, and the PyTorch\footnote{\href{https://pytorch.org/}{PyTorch}} version is 2.0.

\subsection{Datasets and Hyperparameter settings}
 We first conduct experiments with various teacher-student pair settings on the CIFAR-100 dataset~\cite{krizhevsky2009learning}. CIFAR-100 contains 50,000 training images with 500 images per class and 10,000 test images. Except for the loss function, training settings like learning rate or training epochs are the same with Tian et al.~\cite{tian2019contrastive} for CIFAR-100. During each iteration of the training process, for models such as MobileNetV2, ShuffleNetV1, and ShuffleNetV2, we use a learning rate of 0.01 and train for 30 epochs. For other models in CIFAR-100 classification tasks, the learning rate is initialized at 0.1 and the decay rate is 0.1 per 20 epochs, continuing until 60 epochs. The batch size is 64 and the parameter $\alpha$ in Equation \ref{eq:optimize-objective-student} is 0.6. In knowledge distillation process, the teacher model is well-trained previously and fine-tuned during training.

 In the experiments of semi-supervised learning, we consider two standard benchmarks: CIFAR-10-4k and SVHN-1k, which have been widely used in the literature to compare different semi-supervised learning algorithms. These benchmarks were created by keeping a small fraction of the training set as labelled data while using the rest as unlabelled data. For CIFAR-10~\cite{krizhevsky2009learning}, 4,000 labelled examples are kept as labelled data while 41,000 examples are used as unlabelled data. The test set for CIFAR-10 is standard and consists of 10,000 examples. For SVHN~\cite{netzer2011reading}, 1,000 examples are used as labelled data whereas about 603,000 examples are used as unlabelled data. The test set for SVHN is also standard and has 26,032 examples. In the CIFAR-10 classification task, the parameter settings for the supervised and semi-supervised scenarios are exactly the same. The teacher and student models use the same parameter configuration as the corresponding models in the CIFAR-100 task. For the SVHN classification task, the parameters of the supervised and semi-supervised settings also remain unchanged. The initial learning rate of the teacher and student models is set to 0.01, and the decay rate is 0.5 per 20 epochs, continuing until 40 epochs.

In the experiments of Natural Language Processing, we use the GLUE benchmark~\cite{wang2018glue}, which provides a comprehensive suite of tasks. Specifically, we assess our method on the four widely used datasets: MRPC, SST-2, QNLI and RTE. MRPC dataset~\cite{dolan2005automatically} consists of 5,801 sentence pairs, where the task is to determine whether two sentences in a pair are semantically equivalent (paraphrases). SST-2 dataset~\cite{socher2013recursive} is a subset of the Stanford Sentiment Treebank that focuses on binary sentiment classification. The task involves determining whether a given sentence expresses positive or negative sentiment. MNLI dataset~\cite{williams2017broad} consists of 433,000 sentence pairs across multiple genres, where the task is to determine the relationship between a premise and a hypothesis. The RTE dataset~\cite{wang2018glue} includes 2,500 sentence pairs where the task is to decide whether a hypothesis can be inferred (entails) from a given premise. This dataset originates from the Recognizing Textual Entailment challenges. During each iteration of the training process, the maximum sequence length for both teacher and student models is set to 128. The teacher model uses a fixed learning rate of 3e-5 and is fine-tuned for 6 epochs. For the student model, the learning rate varies within $\{1 \times 10^{-4}, 3 \times 10^{-5}, 5 \times 10^{-5}\}$, and it is updated every 6 epochs, continuing until a total of 18 epochs. The batch size is 32 and the parameter $\alpha$ in Equation \ref{eq:optimize-objective-student} is 0.6.

\subsection{Experimental Results and Analysis}
 
We first discuss different update mechanisms involving distillation influence weights. Building on this foundation, we evaluate our KD-AIF framework across four knowledge distillation tasks: (a) applying KD-AIF in image classification tasks; (b) applying KD-AIF in Natural Language Processing tasks; (c) assessing the performance of KD-AIF on datasets containing noisy data; and (d) extending its application to semi-supervised learning scenarios. Through these tasks, we aim to showcase the broad applicability and potential of KD-AIF to significantly improve knowledge distillation outcomes across different domains and challenges.

\subsubsection{Different update mechanism with distillation influence weight}
\begin{figure}[!ht]
    \centering 
    \includegraphics[width=0.9\textwidth]{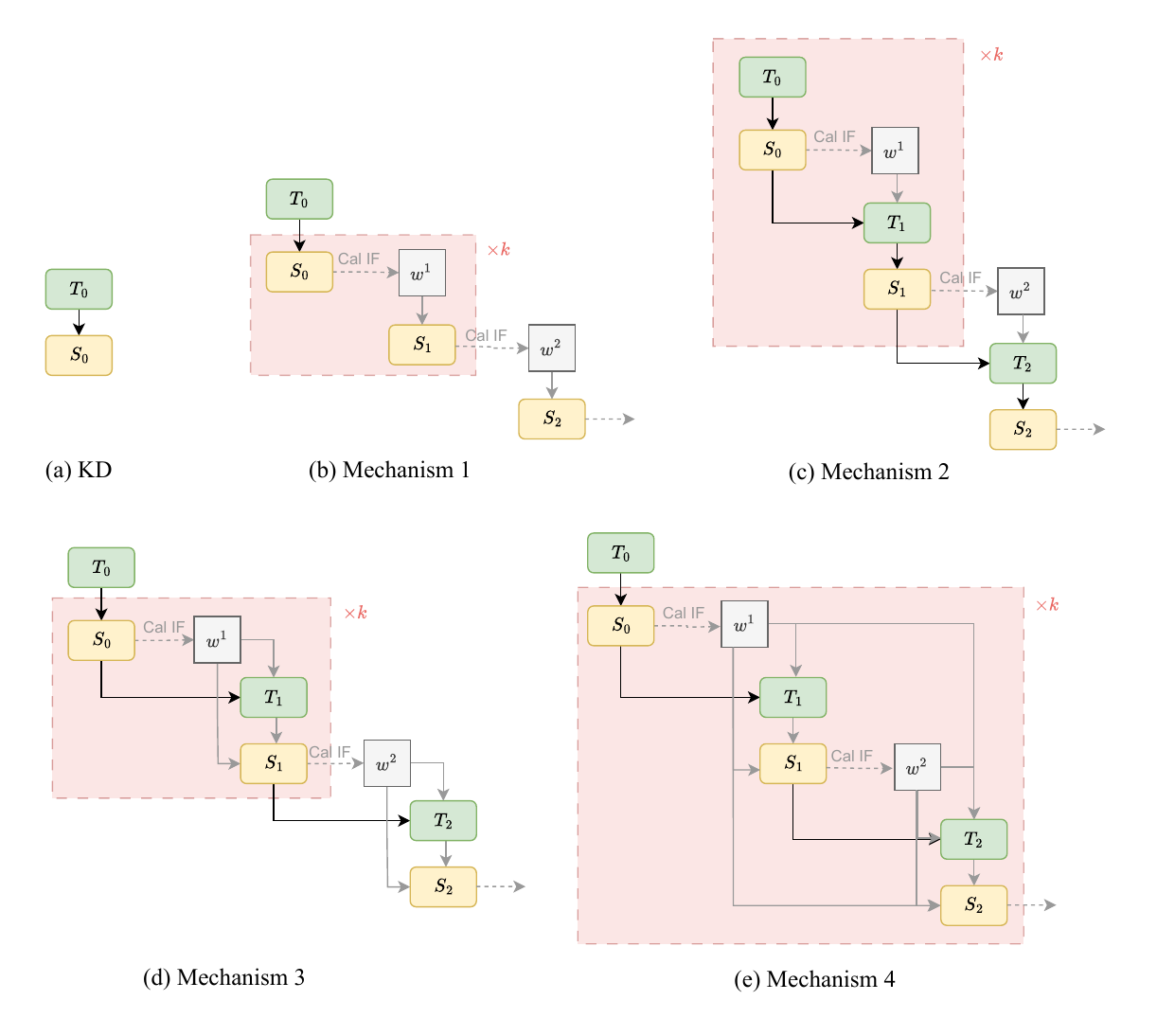}
\caption{Different update mechanisms. $T$ and $S$ denote the teacher model and the student model, respectively. $w$ denotes the influence weight calculated on the student model. The subscript number represents the distillation iteration. In Mechanism 1, the influence weight is used solely to update the student model, while in Mechanism 2, it is used exclusively to update the teacher model. Mechanism 3 proposes using the influence weight for updating both the teacher and student models. Building on Mechanism 3, Mechanism 4 also takes into account the influence weight calculated by the previous student model. Each mechanism will repeat the distillation process $k$ times.}
\label{fig_4}
\end{figure}

We have proposed distillation influence weight to estimate how distilling on each training sample impacts students' performance on the validation set. To seamlessly weave the data influence into the fabric of the teacher-student architecture, we explored and implemented four distinct strategies, as meticulously illustrated in Figure \ref{fig_4}, to integrate influence weight throughout the training regimen. Each mechanism represents a unique approach to integrating influence weight into the knowledge distillation process, offering various strategies for optimizing the interplay between teacher and student models for improved performance and learning efficiency. By conducting a comparison of these four update mechanisms against the foundational knowledge distillation framework initially proposed by Hinton et al.~\cite{hinton2015distilling}, as shown in  Figure \ref{fig_4}(a),  we aim to not only underscore the intrinsic value of incorporating distillation influence weight but also to discern which update strategy for influence weight most aptly complements the demands of the specified task. 

\begin{enumerate}[(1)]
    \item Mechanism 1: In this mechanism, the influence weight is only used to update the student model, focusing solely on enhancing the student's performance based on the distilled knowledge from the teacher without modifying the teacher model itself. This mechanism prioritizes direct improvements in the student model's learning capability from the existing knowledge base, and it can be seen as a variation of traditional offline learning, where the teacher model does not receive any feedback from the student.
    \item Mechanism 2: In this mechanism, the influence weight is only used to update the teacher model, potentially improving the quality of knowledge transferred to the student. By optimizing the teacher's understanding, the aim is to indirectly improve the student's learning outcomes through enhanced guidance.
    \item Mechanism 3: This mechanism advocates for the concurrent update of both teacher and student models using influence weight. By allowing both models to learn and adapt together, this mechanism seeks to create a more dynamic and responsive teaching and learning environment, potentially leading to more significant overall performance improvements.
    \item Mechanism 4: This mechanism not only updates the data weight of both teacher and student models simultaneously but also incorporates the influence weight calculated from the previous student model's performance. This mechanism comprehensively considers all historical performances of the student model on the validation set..
\end{enumerate}

\begin{table}[h]
\setlength{\tabcolsep}{10.2mm}{
\centering
\begin{tabular}{ccccc}
\hline
\multicolumn{1}{c}{\multirow{2}{*}{}}  & \multicolumn{2}{c}{CIFAR-10} & \multicolumn{2}{c}{SVHN} \\ \cline{2-5} 
\multicolumn{1}{c}{}        
& Teacher   & Student   & Teacher  & Student\\ \hline
KD                          
& 0.7344    & 0.5555    & 0.8874   & 0.7392 \\
Mechanism 1                    
& -         & 0.5662    & -        & 0.7529 \\
Mechanism 2                    
& 0.7613    & 0.6000    & \textbf{0.8984}   & 0.7039 \\
Mechanism 3                
& \textbf{0.7640}    & \textbf{0.6068}    & 0.8916   & \textbf{0.7559} \\
Mechanism 4  
& 0.7550    & 0.5960    & 0.8952   & 0.7423 \\ \hline
\end{tabular}
}
\caption{Influence weight on different positions during knowledge distillation training on CIFAR-10 and SVHN dataset. Overall, Mechanism 3 which employs influence weight in the training of both teacher and student models achieves the best performance. For subsequent experiments, we elect to proceed with Mechanism 3 as our comparative benchmark
against other approaches.}
\label{tab_1}
\end{table}

As highlighted in Table \ref{tab_1}, the results of the CIFAR-10 and SVHN datasets reveals that incorporating students' feedback into the training regimen significantly enhances the performance of both teacher and student models. When the distillation process of each mechanism is repeated more than five times, the results tend to be stable, so $k$ in each mechanism is 5. Notably, in the CIFAR-10 dataset, the inclusion of students' feedback yields the most substantial improvement in both models' training processes. However, in the SVHN dataset, the teacher model's performance enhancement from student feedback is somewhat muted, suggesting a greater benefit to the student model itself. This differential impact could be attributed to the teacher model nearing the threshold of overfitting within this particular dataset. Overall, our findings indicate that employing influence weight in the training of both teacher and student models achieves the best performance. Consequently, for subsequent experiments, we elect to proceed with Mechanism 3 as our comparative benchmark against alternative approaches, underscoring its effectiveness in optimizing knowledge distillation processes.

\subsubsection{Comparison with Other Distillation Methods in Image Classification Tasks}
For comparison, the following recent state-of-the-art methods are chosen: FitNet (proposed by Romero et al.~\cite{romero2014fitnets} in 2014), AT (proposed by Zagoruyko \& Komodakis~\cite{zagoruyko2016paying} in 2016), SP (proposed by Tung \& Mori~\cite{tung2019similarity} in 2019), CC (proposed by Peng et al.~\cite{peng2019correlation} in 2019), VID (proposed by Ahn et al.~\cite{ahn2019variational} in 2019), RKD (proposed by Park et al.~\cite{park2019relational} in 2019), PKT (proposed by Passalis \& Tefas~\cite{passalis2018learning} in 2018), AB (proposed by Heo et al.~\cite{heo2019comprehensive} in 2019), FT (proposed by Kim et al.~\cite{kim2018paraphrasing} in 2018), NST (proposed by Huang \& Wang~\cite{huang2017like} in 2017), CRD (proposed by Tian et al.~\cite{tian2019contrastive} in 2019) and CRD+CutMixPick (proposed by Wang et al.~\cite{wang2022makes} in 2022).

\begin{table}[h]
\resizebox{\linewidth}{!}{
\begin{tabular}{ccccccc}
\hline
Teacher       & vgg13   & ResNet50  & ResNet50  & ResNet32*4    & ResNet32*4    & WRN-40-2  \\
Student       & MobileNetV2 & MobileNetV2   & vgg8  & ShuffleNetV1  & ShuffleNetV2  & ShuffleNetV1  \\ \hline
Teacher       & 74.64   & 79.34 & 79.34 & 79.42 & 79.42 & 75.61 \\
Student       & 64.6    & 64.6  & 70.36 & 70.5  & 71.82 & 70.5  \\ \hline
KD            & 67.37   & 67.35 & 73.81 & 74.07 & 74.45 & 74.83 \\
FitNet        & 64.14   & 63.16 & 70.69 & 73.59 & 73.54 & 73.73 \\
AT            & 59.4    & 58.58 & 71.84 & 71.73 & 72.73 & 73.32 \\
SP            & 66.3    & 68.08 & 73.34 & 73.48 & 74.56 & 74.52 \\
CC            & 64.86   & 65.43 & 70.25 & 71.14 & 71.29 & 71.38 \\
VID           & 65.56   & 67.57 & 70.3  & 73.38 & 73.40 & 73.61 \\
RKD           & 64.52   & 64.43 & 71.5  & 72.28 & 73.21 & 72.21 \\
PKT           & 67.13   & 66.52 & 73.01 & 74.1  & 74.69 & 73.89 \\
AB            & 66.06   & 67.2  & 70.65 & 73.55 & 74.31 & 73.34 \\
FT            & 61.78   & 60.99 & 70.29 & 71.75 & 72.50 & 72.03 \\
NST           & 58.16   & 64.96 & 72.28 & 74.12 & 74.68 & 74.89 \\
CRD           & 69.73   & 69.11 & 74.3  & 75.11 & 75.65 & 76.05 \\ 
CRD+CutMixPick & 70.84   & - & 76.2 & - & 78.51 & - \\
\hline
KD-AIF - Teacher 
    & \multicolumn{1}{c}{75.77} & 82.88 & 83.96 
    & \multicolumn{1}{c}{85.19} & 86.07 & 85.97 \\
KD-AIF - Student 
    & \multicolumn{1}{c}{\textbf{73.31}} & \textbf{70.08} & \textbf{76.95} 
    & \multicolumn{1}{c}{\textbf{79.14}} & \textbf{81.35} & \textbf{78.43} \\ \hline

\end{tabular}
}

\caption{Top-1 classification accuracy results on CIFAR-100. Each teacher-student model pair has a similar architectural style. Higher is better. We report our results over 3 repeated runs.}
\label{tab_2}
\end{table}

\begin{table}[h]
\setlength{\tabcolsep}{3.6mm}{
\centering
\begin{tabular}{ccccccc}
\hline
Teacher       & WRN-40-2   & ResNet56  & ResNet110  & ResNet110    & ResNet32*4    & vgg13  \\
Student       & WRN-16-2 & ResNet20   & ResNet20  & ResNet32  & ResNet8*4  & vgg8  \\ \hline
Teacher       & 75.61   & 72.34 & 74.31 & 74.31 & 79.42 & 74.64 \\
Student       & 73.26   & 69.06 & 69.06 & 71.14 & 72.50 & 70.36  \\ \hline
KD            & 74.92   & 70.66 & 70.67 & 73.08 & 73.33 & 72.98 \\
FitNet        & 73.58   & 69.21 & 68.99 & 71.06 & 73.50 & 71.02 \\
AT            & 74.08   & 70.55 & 70.22 & 72.31 & 73.44 & 71.43 \\
SP            & 73.83   & 69.67 & 70.04 & 72.69 & 72.94 & 72.68 \\
CC            & 73.56   & 69.63 & 69.48 & 71.48 & 72.97 & 70.71 \\
VID           & 74.11   & 70.38 & 70.16 & 72.61 & 73.09 & 71.23 \\
RKD           & 73.35   & 69.61 & 69.25 & 71.82 & 71.90 & 71.48 \\
PKT           & 74.54   & 70.34 & 70.25 & 72.61 & 73.64 & 72.88 \\
AB            & 72.50   & 69.47 & 69.53 & 70.98 & 73.17 & 70.94 \\
FT            & 73.25   & 69.84 & 70.22 & 72.37 & 72.86 & 70.58 \\
NST           & 73.68   & 69.6  & 69.53 & 71.96 & 73.30 & 71.53 \\
CRD           & 75.48   & 71.16 & 71.46 & 73.48 & 75.51 & 73.94 \\ 
CRD+CutMixPick & 76.61   & 72.4 & - & - & - & 75.41 \\
\hline
KD-AIF - Teacher 
    & \multicolumn{1}{c}{83.66} & 82.98 & 83.15 
    & \multicolumn{1}{c}{84.33} & 86.56 & 78.43 \\
KD-AIF - Student 
    & \multicolumn{1}{c}{\textbf{78.23}} & \textbf{77.88} & \textbf{78.85} 
    & \multicolumn{1}{c}{\textbf{79.45}} & \textbf{82.20} & \textbf{75.88} \\ \hline

\end{tabular}
}
\caption{Top-1 classification accuracy results on CIFAR-100. Each teacher-student model pair is from different architectures. Higher is better. We report our results over 3 repeated runs.}
\label{tab_3}
\end{table}

In Table \ref{tab_2} and Table \ref{tab_3}, we present the Top-1 classification accuracy of our method and comparison methods. The results of comparison methods are quoted from relevant papers. Table \ref{tab_2} investigates students and teachers of the same architectural style, while Table \ref{tab_3} focuses on students and teachers from different architectures. The results show that our proposed method KD-AIF uniformly outperform all other distillation models, including the original KD. Surprisingly, in the comparison methods, we find that KD works pretty well and none of the other methods consistently outperforms KD on their own. Some methods such as AT and FitNet that require very similar student and teacher architectures perform quite poorly, even underperform the vanilla student model. One of the recent state-of-the-art models, the Contrastive Representation Distillation (CRD) model, utilizes contrastive learning to align the student model's representation relationships with those of the teacher model, thereby enhancing the student's representation capabilities. However, in CRD, the student model's performance is fundamentally limited by that of the teacher model, as the teacher does not undergo further updates based on feedback from the student. Building on this, Wang et al. extended CRD by introducing data augmentation mechanisms to further enhance its performance. In contrast, our proposed KD-AIF framework incorporates influence weights derived from student feedback, resulting in superior performance. Under the KD-AIF framework, teacher models updated using influence weights show performance improvements that surpass the original, unmodified teacher models. This highlights the critical importance of leveraging student feedback to refine the teacher model's training, thereby enriching and advancing the knowledge distillation process.

\subsubsection{Application in Natural Language Processing}

To evaluate our proposed approach in Natural Language Processing (NLP), we conducted experiments on the GLUE benchmark, which includes a variety of tasks. For comparison, the following recent state-of-the-art methods are chosen: PKD (proposed by Sun et al.~\cite{sun2019patient} in 2019), SKD (proposed by Guo et al.~\cite{guo2020reducing} in 2020), TAKD (proposed by Mizadeh et al.~\cite{mirzadeh2020improved} in 2020), DIST (Huang et al.~\cite{huang2022knowledge} in 2022), PEST-KD (proposed by Rao et al.~\cite{rao2023parameter} in 2023). Our primary objective was to distill knowledge from BERT-Base proposed by Devlin et al.~\cite{devlin2018bert} into a smaller 6-layer BERT model with a hidden size of 768.

We present the performance of our proposed method on the test set of text classification tasks from the GLUE benchmark. As shown in Table \ref{tab_6}, KD-AIF surpasses all baseline methods, including recent competitive knowledge distillation models such as SKD, DIST, and PESF-KD, demonstrating the effectiveness of our approach. Notably, our method achieves state-of-the-art results compared to approaches that rely on carefully crafted training strategies or loss functions, such as PKD, TKAD, and DIST model. PKD employs two distinct distillation schemes that allow the student to extract knowledge incrementally by learning from multiple intermediate layers of the teacher model. TAKD introduces intermediate teacher assistant models to bridge the gap between the teacher and student models, enabling the student model to align more closely with the teacher model. DIST modifies the KL-divergence loss to better align the teacher and student models. Unlike above models, KD-AIF does not depend on a series of teacher assistant models, simplifying the overall framework while maintaining superior performance. Compared to online distillation methods, KD-AIF performs better than SKD and PESF-KD models. This highlights the importance of incorporating student’s feedback during the training process.

\begin{table}[]
\centering
\begin{tabularx}{0.75\textwidth}{l>{\centering\arraybackslash}X>{\centering\arraybackslash}X>{\centering\arraybackslash}X>{\centering\arraybackslash}X}
\hline
Model          & MRPC & RTE  & SST-2 & QNLI \\ \hline
Teacher        & 84.8 & 66.4 & 93.0  & 90.5 \\
Student        & 82.3 & 62.3 & 92.3  & 88.8 \\ \hline
KD             & 81.4 & 64.7 & 91.2  & 89.0 \\
PKD            & 79.9 & 65.5 & 92.0  & 89.0 \\
SKD            & 78.4 & 65.1 & 92.2  & 87.2 \\
TAKD           & 81.7 & 64.1 & 92.5  & 89.4 \\
DIST           & 79.8 & 65.0 & 90.9  & 88.0 \\
PESF-KD        & 80.6 & 65.1 & 91.5  & 87.6 \\ \hline
KD-AIF-Teacher & 86.0 & 67.9 & 93.1  & 91.2 \\
KD-AIF-Student & \textbf{82.8} & \textbf{65.7} & \textbf{92.7}  & \textbf{89.8} \\ \hline
\end{tabularx}
\caption{Performance comparison of various knowledge distillation models on the test sets of the GLUE benchmark for classification tasks.}
\label{tab_6}
\end{table}

\subsubsection{Label Noise Experiments}

In terms of fairness, we acknowledge the work of Chen et al.~\cite{chen2024measuring} on measuring fairness in credit ratings. They evaluate fairness by measuring the rank differences in predictions when specific group variables are removed. Similarly,  in order to evaluate our method in more complex scenarios, we intentionally introduce mislabeled training data to simulate imbalanced and noisy conditions, Table \ref{tab_4} summarizes the performance of our approach under different noise settings on the image classification dataset CIFAR-100 and the text classification dataset SST-2. For the CIFAR-100 classification task, we used a pre-trained ResNet-50 as the teacher model to guide the training of a MobileNetV2 student model. For the SST-2 binary sentiment classification task, we employed a pre-trained BERT-Base teacher model to guide a BERT-6L student model. In this evaluation, we analyzed the importance of incorporating influence weights during distillation.

To introduce noise into the training data, we deliberately mislabelled 10\% and 20\% of the datasets by randomly flipping the labels of the specified proportion of training data to other class labels. This approach introduced a bias in the originally balanced training datasets, resulting in some classes having fewer noisy data while others contained more. Even under these challenging conditions, our method demonstrated that distillation with influence weights consistently outperformed models using uniform weights across noise settings with different noise ratios.

\begin{figure}[!ht]
    \centering 
    \includegraphics[width=\textwidth]{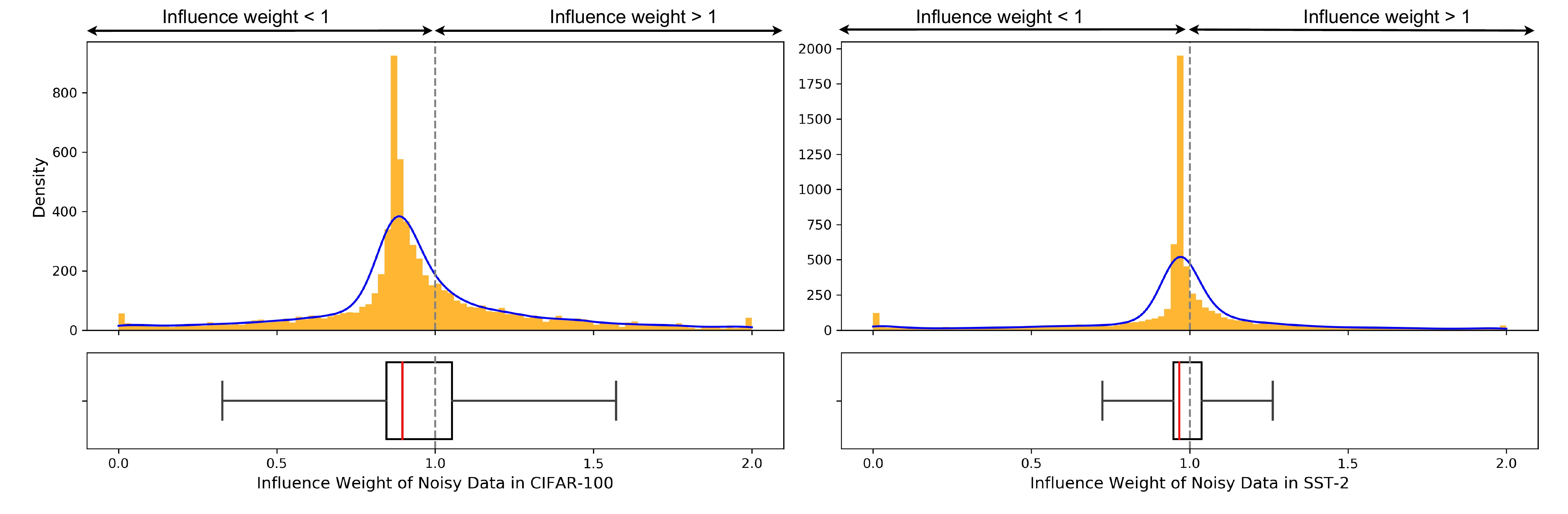}
\caption{Distribution of influence weights on 10\% noisy training data in the CIFAR-100 and SST-2 datasets. If the influence weight is less than 1, the corresponding data is considered harmful. Our proposed method KD-AIF demonstrates the ability to accurately identify a majority of noisy data across different classification tasks.}
\label{fig_5}
\end{figure}

To better understand why influence weights contribute to learning more robust models during training, we visualized the weight distribution of noisy training data with a 10\% noise ratio for both classification tasks in Figure \ref{fig_5}. The results reveal that most noisy training data received influence weights below 1, effectively identifying them as harmful data. This finding strongly supports the effectiveness of our model in accurately identifying noisy training data across different tasks, lowering their weights, and thus ensuring a more efficient and reliable training process.

\begin{table}[]
\setlength{\tabcolsep}{6.65mm}{
\begin{tabular}{cccccc}
\hline
\multirow{2}{*}{Dataset}  & \multirow{2}{*}{Teacher Model} & \multirow{2}{*}{Student Model} & \multirow{2}{*}{Method} & \multicolumn{2}{c}{Noise Rate} \\ \cline{5-6} 
                          &                                &                                &                         & 10\%           & 20\%          \\ \hline
\multirow{2}{*}{CIFAR-100} & \multirow{2}{*}{Resnet 50}     & \multirow{2}{*}{MobleNetV2}    & KD                      & 66.05     & 57.86    \\
                          &                                &                                & KD-AIF                  & \textbf{69.61}     & \textbf{59.25}    \\
\multirow{2}{*}{SST-2}    & \multirow{2}{*}{Bert-Base}     & \multirow{2}{*}{Bert-6L}       & KD                      & 88.67     & 85.86    \\
                          &                                &                                & KD-AIF                  & \textbf{90.58}    & \textbf{89.44}    \\ \hline
\end{tabular}
}
\caption{Top-1 classification accuracy results on CIFAR-100 dataset and SST-2 dataset under different noise rates. Noise rate is the ratio that we intentionally mislabelled in the training set.}
\label{tab_4}
\end{table}

\subsubsection{Application in Semi-Supervised Learning}
We use WideResNet-28-2 as the teacher model which has 1.45 million parameters and select MobileNetV2 as the student model. To ensure a fair comparison, we only compare KD-AIF against methods that use the same architectures. The baseline methods include Temporal Ensemble (proposed by Laine and Aila~\cite{laine2016temporal} in 2016), Mean Teacher (proposed by Tarvainen and Valpola~\cite{tarvainen2017mean} in 2017), VAT+EntMin (proposed by Miyato et al.~\cite{miyato2018virtual} in 2018), LGA+VAT (proposed by Jackson and Schulman~\cite{jackson2019semi} in 2019), ICT (proposed by Verma et al.~\cite{verma2022interpolation} in 2022). During the training phase where we train both the teacher and the student, we use the default hyper-parameters from previous work for all our models. 

Table \ref{tab_5} presents our results with KD-AIF-Teacher in comparison with other methods. In semi-supervised learning tasks, we use the teacher model with distillation influence weight to compare with other baselines since most of the semi-supervised methods only contain a single model and the role of the student model is to assist in obtaining a better teacher model. The results show that under strictly fair comparisons, our method outperforms other baselines. In semi-supervised learning, at this level of accuracy, our gain of 1\%-2\% over ICT methods~\cite{verma2022interpolation} is a very significant margin of improvement. If we consider adding data enhancement techniques to enhance the pseudo-labelling capability in this task, our method still has room for further performance enhancement.

\begin{table}[]
\centering
\begin{tabularx}{0.75\textwidth}{l>{\centering\arraybackslash}X>{\centering\arraybackslash}X}
\hline
\multicolumn{1}{c}{Semi-Supervised Learning Method} & CIFAR-10-4k    & SVHN-1k    \\ \hline
Temporal Ensemble              & 83.63          & 92.81      \\
Mean Teacher                   & 84.13          & 94.35      \\
VAT+EntMin                     & 86.87          & 94.65      \\
LGA+VAT                        & 87.94          & 93.42      \\
ICT                            & 92.17          & 96.11      \\ \hline
KD-AIF-Teacher           & \textbf{92.35} & \textbf{96.21} \\
\hline
\end{tabularx}
\caption{Image classification accuracy of semi-supervised methods on CIFAR-10-4k and SVHN-1k. For a fair comparison, we only include results that share the same model architecture WideResNet-28-2 for both datasets.}
\label{tab_5}
\end{table}

\section{Conclusion}\label{sec:5}
\chreplaced{}{}In this paper, we first revisit several teacher-student architectures utilized in knowledge distillation. We then introduce the concept of distillation influence, which quantifies how the distillation process from each training data affects the student model's generalization ability. Building on this concept and following the SAFE model concept, we propose a systematic framework called KD-AIF (Knowledge Distillation with Adaptive Influence Weight). In this paper, we revisited several teacher-student architectures commonly employed in knowledge distillation. We introduce the novel concept of distillation influence, which quantifies the impact of the distillation process from individual training data points on the student model's generalization performance. Leveraging this concept, and drawing inspiration from the SAFE model concept, we proposed a systematic approach called KD-AIF (Knowledge Distillation with Adaptive Influence Weight), offering a more nuanced and effective way to enhance the student model's learning process.

The learning process within KD-AIF comprises two main updates: first, adjusting the teacher model based on the student model's performance on unseen data; second, updating the student model using the soft labels generated by the teacher model along with the previous scoring of the data by the student model. Theoretically, we demonstrate that incorporating influence weights leads to a model with improved generalization capabilities. In experiments on both image classification and text classification tasks, we validate the superiority of the KD-AIF model and its applicability across different domains. This also highlights the importance of feedback from the student model based on its performance on unseen datasets.

When applying KD-AIF to semi-supervised learning, the core idea is that the teacher model learns from the student's feedback to generate pseudo-labels that optimally facilitate the student’s learning process. Experiments conducted on standard low-resource benchmarks, such as CIFAR-10-4k and SVHN-1k, demonstrate that KD-AIF outperforms many existing semi-supervised learning methods. This highlights that KD-AIF not only enhances knowledge distillation but also provides significant improvements in scenarios with limited labelled data, further establishing its versatility and robustness in various learning contexts.

Overall, the proposed KD-AIF framework bridges key concepts such as sustainability(robustness), accuracy, fairness, and explainability with the established knowledge distillation model. By dynamically adjusting training sample weights, the framework enhances model robustness against noisy labels and distribution shifts, ensuring sustainable performance in resource-constrained scenarios. What's more, its prioritization of informative data improves accuracy while aligning with fairness principles by mitigating biases or mislabelling in training data. Moreover, the influence function introduces the interpretability of the black-box model, enabling researchers to trace model decisions back to individual data points, thereby supporting transparency and informed optimization in AI systems. Despite these advantages, the current algorithm still has some limitations, as it is rarely used in large models due to the computational challenges associated with calculating the inverse Hessian matrix. Consequently, this paper adopts a student model with a simplified framework to compute the influence score. Given the complexity of the influence function, exploring theoretical approximations represents a promising direction for enabling its application to more complex models, such as LLMs. This work represents an organic integration of statistical methods and AI, leveraging the rigorous interpretability and theoretical foundations of influence functions from statistics to enhance the performance and robustness of AI models. As part of the future work, we aim to develop scalable approximation methods for the influence function that maintain accuracy while reducing computational overhead. Additionally, we plan to extend this framework to address specific challenges in large-scale applications, such as handling highly imbalanced data or incorporating multimodal learning scenarios. What's more, by leveraging the underlying concordance metrics of the Rank Graduation Box (RGB), we could systematically measure robustness to extreme conditions, precision of predictions, equitable treatment across subgroups of datasets, and interpretability of model outputs. This integration strengthens the evaluation framework by aligning the KD-AIF with robust statistical metrics to ensure accountability and transparency.

\section{Disclosure statement}
No potential conflict of interest was reported by the authors.

\section{Acknowledgement}
This work is partially supported by Guangdong Provincial Key Laboratory of Interdisciplinary Research and Application for Data Science, BNU-HKBU United International College, project code 2022B1212010006, UIC research grant R0400001-22; National Natural Science Foundation of China (No.12231004), UIC research grant UICR0600048; National Natural Science Foundation of China (No.1272054), UIC research grant UICR0600036; Guangdong University Innovation and Enhancement Programme Funds Featured Innovation Project 2018KTSCX278, UIC Research Grants R5201910; UIC Research Grants R201809; European Research Council (ERC) under the European Union's Horizon 2020 research and innovation programme (grant agreement No. 101002240). The computations in this paper were performed using the Bayes Cluster (USBC) provided by the Department of Statistics and Data Science, BNU-HKBU United International College.

\bibliographystyle{unsrt}  
\bibliography{reference}  

\section*{Appendix}
\renewcommand{\theequation}{A\arabic{equation}}
\setcounter{equation}{0}
\setcounter{Theorem}{1}
\subsection*{A. Proof of Theorem 1}\label{appendix:theorem-1}
\begin{proof}
    Let $\epsilon_{i} = f_{\epsilon}(\phi_i(\hat{\theta}_S))$, $f_{\epsilon} \in \mathcal{F}:\mathbb{R} \to \mathbb{R}$. Based on the two conditions mentioned above, it is clear that we have $f_{\epsilon}(0) = 0 $, $f_{\epsilon}(\phi_i(\hat{\theta}_S))>0$ when $\phi_i(\hat{\theta}_S)<0$ and $f_{\epsilon}(\phi_i(\hat{\theta}_S))<0$ when $\phi_i(\hat{\theta}_S)>0$. Each element-wise multiplication of these two vectors $\mathbf{\epsilon}$ and $\Phi(\hat{\theta}_S)$ results in a value less than 0. Therefore, we can draw the conclusion based on this property that $\mathbb{E}_P[\mathbf{\epsilon}\, \Phi(\hat{\theta}_S) ]<0$. 
\end{proof}

\subsection*{B. Proof of Theorem 2}\label{appendix:theorem-2}

\begin{proof}
    
    The worst-case risk $ \acute{R}_{\hat{\theta}_\epsilon}(Q) $ is defined as the supremum of the risk over all $ Q \in \mathcal{Q} $
\begin{equation}
    \acute{R}_{\hat{\theta}_{S,\epsilon}}(Q) = \sup_{Q \in \mathcal{Q}} \mathbb{E}_{Q} \left[ L_{S}(z^{val};\hat{\theta}_{S,\epsilon},\theta_T)  \right]
\end{equation}

The dual representation of this equation is given by:
\begin{equation}\label{eq:appendix}
\acute{R}_{\hat{\theta}_{S,\epsilon}}(\eta) = \inf_{\eta \in \mathbb{R}} \left( \sqrt{2\delta + 1} \times \mathbb{E}_P \left[ (L_{S}(z^{train};\hat{\theta}_{S,\epsilon},\theta_T) - \eta)_+^2 \right]^{\frac{1}{2}} + \eta \right)
\end{equation}

The gradient of $\acute{R}_{\hat{\theta}_{S,\epsilon}}(\eta)$ on the influence function vector $ \Phi(\hat{\theta}_S) $ is a vector:
\begin{equation}
    \nabla_{\boldsymbol{\phi}} \acute{R}_{\hat{\theta}_{S,\epsilon}}(\hat{\eta}) 
    = (\frac{\partial \acute{R}_{\hat{\theta}_{S,\epsilon}}(\hat{\eta}) }{\partial  \phi_1(\hat{\theta}_S)} ,
    \frac{\partial \acute{R}_{\hat{\theta}_{S,\epsilon}}(\hat{\eta}) }{\partial  \phi_2(\hat{\theta}_S)},
    \cdots,
    \frac{\partial \acute{R}_{\hat{\theta}_{S,\epsilon}}(\hat{\eta}) }{\partial  \phi_{N_{train}}(\hat{\theta}_S)}
    )^{T}
\end{equation}
where $\hat{\eta}$ can reach the infimum in Equation \ref{eq:appendix}. Take one element of the gradient and analyze the bound:

\begin{equation}
    \begin{aligned}
        \frac{\partial \acute{R}_{\hat{\theta}_{S,\epsilon}}(\hat{\eta}) }{\partial  \phi_i(\hat{\theta}_S)}
        &= \sqrt{2\delta + 1} \times \frac{\partial\mathbb{E}_P \left[ (L_{S}(z^{train};\hat{\theta}_{S,\epsilon},\theta_T) - \hat{\eta})_+^2 \right]^{\frac{1}{2}}}{\partial  \phi_i(\hat{\theta}_S)}\\
        &= \sqrt{2\delta + 1} \times \frac{\partial \left[ (L_{S}(z_i^{train};\hat{\theta}_{S,\epsilon},\theta_T) - \hat{\eta})_+^2 \right]^{\frac{1}{2}}}{\partial  \phi_i(\hat{\theta}_S)}\\
        &\leq  \sqrt{2\delta + 1} \times \frac{\partial |L_{S}(z_i^{train};\hat{\theta}_{S,\epsilon},\theta_T) - \hat{\eta}| }{\partial  \phi_i(\hat{\theta}_S)} \\
        &\leq  \sqrt{2\delta + 1} \times |\frac{\partial L_{S}(z_i^{train};\hat{\theta}_{S,\epsilon},\theta_T) }{\partial  \epsilon_i} |\times |\frac{\partial  \epsilon_i}{\partial  \phi_i(\hat{\theta}_S)}| \\
        &\leq  \sqrt{2\delta + 1} \times |\frac{\partial (L_{S}(z_i^{train};\hat{\theta}_{S,\epsilon},\theta_T) -L_{S}(z_i^{train};\hat{\theta}_{S},\theta_T)) }{\partial  \epsilon_i} |\times |\frac{\partial  \epsilon_i}{\partial  \phi_i(\hat{\theta}_S)}| \\
        &\leq  \sqrt{2\delta + 1} \times \frac{\sigma}{N_{train}} \times |\phi_i(\hat{\theta}_S)|      
    \end{aligned}
\end{equation}

Hence we can get the conclusion
\begin{equation}
  \|\nabla_{\boldsymbol{\phi}} \acute{R}_{\hat{\theta}_{S,\epsilon}}(\hat{\eta})\|
  \leq \frac{\sigma\sqrt{2\delta + 1}}{N_{train}}\times \|\Phi(\hat{\theta}_S)\|
\end{equation}

That means the change rate of $\acute{R}_{\hat{\theta}_{S,\epsilon}}(\hat{\eta})$ is aligned with $\xi =\mathcal{O}( \frac{\sigma\sqrt{2\delta + 1}}{N_{train}}) $.

\end{proof}

\section*{Author Details}
\textbf{Sirong Wu} is currently pursuing the PhD degree in the Department of Data Science and Statistics at Beijing Normal University-Hong Kong Baptist University United International College. Her research interests include deep learning, and explainable artificial intelligence. 

\textbf{Xi Luo} is currently pursuing the PhD degree in the Department of Data Science and Statistics at Beijing Normal University-Hong Kong Baptist University United International College. Her research interests are deep learning, medical image weakly supervised segmentation, and explainable artificial intelligence.

\textbf{Junjie Liu} holds an MPhil degree in Probability and Mathematical Statistics from Hong Kong Baptist University, and his research interests include statistical inference, spatio-temporal models, and text analysis. He is currently a PhD student at Trinity College Dublin and investigating conflict forecasting.

\textbf{Yuhui Deng}  is currently a professor in the Department of Statistics and Data Science at the Faculty of Science and Technology of the Beijing Normal University-Hong Kong Baptist University United International College (UIC). He obtained his Bachelor's degree in Applied Mathematics from Wuhan University, a Master's degree in Probability Theory and Mathematical Statistics, and a Ph.D. in Applied Mathematics from City University of Hong Kong. Prof. Deng is currently a committee member of the Uniform Design Association of China, and a member of China Society for Industrial and Applied Mathematics. Prof. Deng’s research interests include data mining, machine learning, deep learning, and experimental design. He has published numerous related papers in international academic journals and conferences and has led multiple research funding projects. In 2018, he was awarded the President's Award for Teaching and Service at UIC. In 2021, he was honoured with the title of "Outstanding Teacher in Nan Yue" by the Guangdong Provincial Department of Education.

\end{document}